\documentclass{article}
\usepackage{iclr2026_conference}

\iclrfinalcopy   

\usepackage{times}
\usepackage{latexsym}
\usepackage[T1]{fontenc}
\usepackage[utf8]{inputenc}
\usepackage{microtype}     
\usepackage{inconsolata}   
\usepackage{graphicx}

\usepackage{amsmath}
\usepackage{subcaption}    
\usepackage{algorithm}     
\usepackage{algpseudocode} 
\usepackage{enumitem}
\usepackage{pifont}
\usepackage{booktabs}

\usepackage{xcolor}
\definecolor{deepblue}{RGB}{0, 35, 102}
\usepackage[colorlinks=true, linkcolor=deepblue, citecolor=deepblue, urlcolor=deepblue]{hyperref}

\newcommand{\stitle}[1]{\vspace*{0.4em}\noindent{\bf #1.\/}}

\newcommand{\squishlist}{
	\begin{list}{$\bullet$}
		{ \setlength{\itemsep}{1pt}
			\setlength{\parsep}{1pt}
			\setlength{\topsep}{2.5pt}
			\setlength{\partopsep}{0.5pt}
			\setlength{\leftmargin}{1em}
			\setlength{\labelwidth}{1em}
			\setlength{\labelsep}{0.6em}
		}
	}
	\newcommand{\squishend}{
	\end{list}
}

\title{Metis: Bridging Text and Code Memory for Self-Evolving Agents
}

\author{%
\begin{minipage}{\textwidth}
\centering
\vspace{6pt}%
{\bf Zijie Dai$^{1}$\quad Siuhin He$^{1}$\quad Hui Li$^{1}$\quad Qihui Zhou$^{2}$\quad Jiajun Li$^{2}$\quad Mingcong Song$^{2}$}\\[3pt]
{\bf Guoping Long$^{2}$\quad Hongjie Si$^{2}$\quad Xin Yao$^{2}$\quad Lin Zhang$^{2}$\quad James Cheng$^{1}$\quad Xiao Yan$^{3}$}\\[7pt]
{\normalfont $^{1}$The Chinese University of Hong Kong\qquad
$^{2}$Huawei\qquad
$^{3}$Wuhan University}\\[3pt]
{\normalfont\texttt{caiusdai@link.cuhk.edu.hk}}
\end{minipage}
}

\begin{document}
\maketitle

\begin{abstract}
Self-evolving agents improve over time by distilling experience from past executions and reusing it in future tasks.
Existing systems represent such experience either as natural-language text injected into the agent context or as code exposed as callable tools. However, the choice between these representations is typically made at design time rather than derived from the characteristics of the experience itself, leaving the trade-offs between them poorly understood.
We present the first controlled study that isolates text memory and code memory over an identical set of experiences. Our results show that the two forms exhibit complementary trade-offs in construction cost, execution efficiency, and transferability, such that neither representation alone is sufficient.
Guided by these findings, we propose Metis, a self-evolving agent system built on a hierarchical dual-representation memory. Metis organizes textual experience into execution plans, environment facts, and common pitfalls, and selectively crystallizes recurring plans into validated callable tools. This design combines the broad applicability of text memory with the execution efficiency of code memory while incurring tool-generation cost only when justified by repeated reuse.
We evaluate Metis on AppWorld, a challenging benchmark for interactive agents. The results show that Metis improves task accuracy by up to 20.6\% over ReAct while reducing execution cost by up to 22.8\%. Compared with representative self-evolving agent systems, Metis consistently achieves a better balance between accuracy, execution efficiency, and memory-construction cost.

\end{abstract}

\section{Introduction}
\label{sec:introduction}

Recent advances in large language models (LLMs) have driven a shift from passive text generation toward agentic systems \citep{llm_to_agents} that interleave reasoning with action \citep{react} and invoke external tools \citep{tooluse,mcp} to accomplish user-specified tasks.
Once deployed, however, agents operate in open-ended environments whose challenges cannot be fully anticipated during training. Effective performance therefore requires continual adaptation through interaction with the environment rather than relying solely on capabilities acquired at training time \citep{evolve_survey_questions}.
A fundamental obstacle to such adaptation is the stateless nature of current LLM-based agents. While agents may accumulate valuable experience during task execution, this experience is only retained within the finite context window. Once the interaction history is no longer accessible, the agent loses task-relevant knowledge acquired from previous attempts: when facing related
tasks, it must rediscover solutions it has already found and is prone to repeating mistakes it has already made. Such repetition not only incurs additional computational cost but also degrades task success rates, ultimately limiting the agent's ability to improve over time.

\begin{figure}[t]
\centering
\begin{subfigure}{0.5\textwidth}
  \centering
  \includegraphics[width=\linewidth]{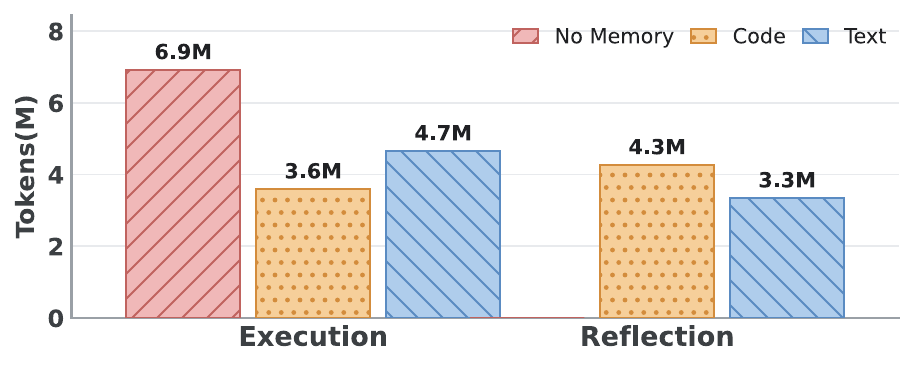}
  \caption{Cost Breakdown}
  \label{fig:exec-refl-token}
\end{subfigure}
\hfill
\begin{subfigure}{0.24\textwidth}
  \centering
  \includegraphics[width=\linewidth]{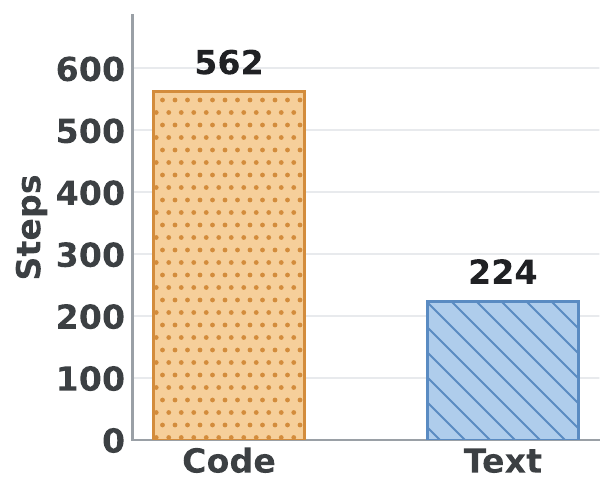}
  \caption{Reflection Steps}
  \label{fig:reflection-turns}
\end{subfigure}
\hfill
\begin{subfigure}{0.24\textwidth}
  \centering
  \includegraphics[width=\linewidth]{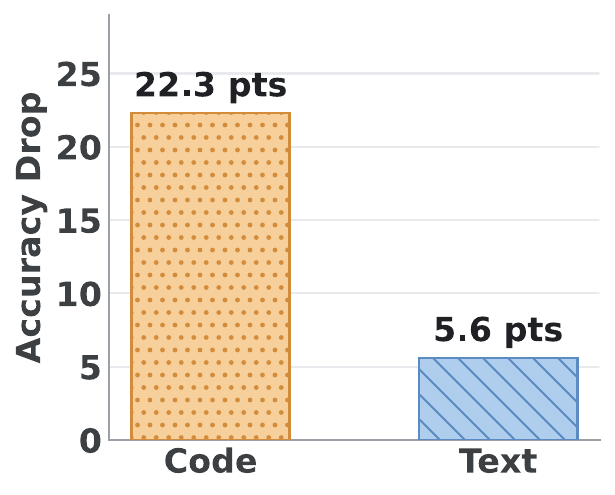}
  \caption{Transfer Reliability}
  \label{fig:transfer-reliability}
\end{subfigure}
\caption{Profiling results of two experience forms.}
\label{fig:trade-offs}
\end{figure}

Self-evolving agents have emerged to address this limitation by distilling experiences from past executions and reusing them in future tasks, enabling continual improvement through interaction with the environment \citep{ace,reme,yunjue,evolverl}.
Existing approaches represent such experiences primarily in two forms, which we distinguish by how the agent consumes the stored experience rather than by its surface content.
The first is \textit{text memory}, where experiences are stored as natural-language knowledge and injected into the agent context at runtime, requiring the agent to read and reason over them \citep{apc,legomem,procmem,dc,ace,memento}.
The second is \textit{code memory}, where past routines are encapsulated as callable tools or MCP services that the agent invokes directly without re-deriving the underlying procedure \citep{yunjue,alita}.
Under this criterion, systems that store code snippets but deliver them through context injection \citep{skillx,xskill} fall into the text-memory category, as the agent still consumes them by reading and reasoning rather than by invocation.

Despite their growing adoption, memory representation remains largely underexplored. Which representation to use, and how the two should cooperate when neither alone suffices, remains largely determined by ad hoc design choices. The field still lacks a systematic understanding of what kinds of experience different representations can capture, what costs they incur, and what unique features they have.

To fill this gap, we conduct a controlled study on the AppWorld \citep{appworld} training split that isolates each representation form over an identical set of experiences. A detailed description of the profiling experiments is provided in Appendix~\ref{app:experiments}. The results, summarized in Figure~\ref{fig:trade-offs}, reveal that text memory and code memory are complementary along three dimensions.

\textit{\textbf{1) Construction Cost.}}
Text memory is substantially cheaper to construct than code memory, as shown in Figure~\ref{fig:exec-refl-token}. Constructing a reusable tool requires additional exploration, validation, and debugging to ensure that the synthesized implementation behaves correctly, incurring roughly $2.5\times$ more ReAct turns as shown in Figure~\ref{fig:reflection-turns} and on the order of one million additional tokens compared with text-memory construction.

\textit{\textbf{2) Execution Efficiency.}}
Code memory affords markedly higher runtime efficiency than text memory as shown in Figure~\ref{fig:exec-refl-token}. A text memory describes a routine in natural language; although such guidance can improve efficiency, the agent must still reason over it and execute the prescribed actions step by step at inference time. In contrast, a code memory compiles the routine into a callable tool, transforming a multi-step reasoning process into a single action invocation and eliminating redundant intermediate reasoning.

\textit{\textbf{3) Transfer Reliability.}}
Text memory generalizes more reliably to unseen tasks than code memory.
We measure this by contrasting two evaluation settings. In the \textit{in-sample} setting, memory is constructed from all tasks and evaluated on the same tasks. In the \textit{streaming} setting, each task is solved using only memory distilled from strictly earlier tasks. The accuracy gap between the two settings therefore, quantifies how much of a representation's benefit fails to transfer beyond the tasks from which it was distilled.
As shown in Figure~\ref{fig:transfer-reliability}, code memory degrades by 22.3 points in the streaming setting, whereas text memory degrades by only 5.6 points. The difference arises from how the two representations are used. Code memories encode experience as fixed executable behavior, which is efficient but brittle under task variation. Text memories instead provide adaptable natural-language guidance that the agent can reinterpret during reasoning, enabling more reliable transfer across related tasks.
Codification should therefore be reserved for experiences exhibiting stable and recurring procedural patterns rather than being applied indiscriminately.

These findings suggest no single memory representation is sufficient. Text memory is cheap to construct and robust under transfer, making it suitable for broadly capturing reusable experience. Code memory, in contrast, offers higher execution efficiency but is costly to build and brittle when transferred to different tasks. Therefore, codification should not be applied indiscriminately. It is only worthwhile when an experience recurs often enough to amortize construction cost and exhibits a stable execution pattern that can be safely crystallized into code. This motivates a selective promotion mechanism, where text memory first serves as a low-cost staging layer, and only recurrent, high-value, and stable experiences are promoted into code.

We thus propose Metis, a hierarchical dual-representation memory system grounded in the findings of our study. At the foundation is text memory, which captures experiences in natural language and is injected at runtime to guide the agent's reasoning. The profiling study also shows that useful textual experience is not homogeneous. Experiences worth recording differ not only in content, but also in the granularity at which they should be stored: some are coarse procedural plans that summarize reusable execution routines, some are fine-grained environment facts that reveal the unique constraints of the environment, and others are localized pitfall warnings that prevent repeated mistakes. Treating all of them as a single flat text tip either over-generalizes task-specific constraints or over-specifies reusable routines. Metis therefore organizes text memory into three categories: plans, facts, and pitfalls. Each category serves a distinct role and is retrieved according to its own scope and usage pattern.

Built on top of text memory is code memory, which serves as a crystallized form of experience. Rather than indiscriminately converting all experiences into code, Metis promotes only recurring plans into callable tools. Facts and pitfalls remain in textual form, as they primarily provide reasoning guidance rather than executable procedures. This selective promotion strategy ensures that code memory is created only when a behavioral pattern has demonstrated sufficient value and stability, amortizing tool-construction cost across repeated use while producing more robust and general tools distilled from multiple observations.

Notably, reflection in Metis lies entirely off the execution critical path: both reflection paths operate on completed trajectories and can thus run asynchronously in real-world deployments, so memory construction never inflates task-serving latency. By combining structured textual guidance with selectively crystallized executable knowledge, Metis preserves the low construction cost and high transfer reliability of text memory while leveraging the execution efficiency of code memory. To summarize, this paper makes the following contributions:

\squishlist
    \item We present, to the best of our knowledge, the first controlled study of experience representation that isolates memory representation forms over an identical set of experiences and a shared agent backbone, disentangling the effect of representation from that of the underlying agent system.
    
    \item We introduce Metis, a hierarchical dual-representation memory system grounded in the insights of our representation study. Metis treats text memory as the foundation and selectively promotes recurring strategic plans into callable code, while retaining environment facts and common pitfalls as textual guidance.
    
    \item We demonstrate through extensive experiments on AppWorld benchmark~\citep{appworld} that Metis substantially improves both execution efficiency and task success rate over existing memory-based agents, while keeping memory-construction cost comparable to or lower than prior self-evolving approaches.
\squishend

\section{Preliminaries}
\label{sec:preliminaries}
\subsection{Agentic Workflow} \label{sec:prelim-agent}
We consider a paradigm that is commonly used in the industry, where an agent solves a user task by interacting with a stateful sandbox environment through ReAct loop \citep{react,tau-bench,sweagent}.
A task query given by the user is a natural-language instruction $q$, and the agent is an LLM policy $\pi$ acting over an interaction context $c_t$ ($c_1$ holds $q$ and a system prompt). 
At step $t$ the policy produces an action $a_t \sim \pi(\cdot \mid c_t)$, either a code block or a tool call, treated uniformly since both are text the system parses then executes \citep{codeagent, tooluse}, and the environment returns an observation $o_t$, extending the context as $c_{t+1} = c_t \oplus (a_t, o_t)$.
The interaction halts at step $T$, either at a terminal action or a step budget, producing a trajectory: $\tau = \big( q,\; (a_1, o_1),\; \dots,\; (a_T, o_T) \big).$

\subsection{Self-Evolving Agent Systems} \label{sec:prelim-evolve}
Self-evolving agents improve over time by incorporating experience acquired from past task executions. Existing approaches can be broadly categorized into two paradigms. One updates the policy parameters through reinforcement learning on collected experience \citep{absolutezero,evolverl}. Such approaches internalize acquired experience into model parameters, making it difficult to inspect, transfer across model backbones, or reuse when the underlying model is replaced \citep{trajectory_rl}. In this work, we focus on the complementary \emph{training-free} paradigm, in which experience is retained as an explicit, model-agnostic memory artifact that conditions a frozen policy during inference.

Concretely, the agent faces a stream of tasks $q^{(1)}, q^{(2)}, \dots$ and maintains a memory $M$, initially empty ($M_0 = \emptyset$). For task $q^{(k)}$, relevant entries are retrieved from the current memory $M_{k-1}$ , usually by applying similarity-based vector search techniques \citep{vs,legomem,vs2} and LLM-based filtering \citep{yunjue,reme}, which are then incorporated into the interaction context. The policy is then conditioned on the retrieved memory, $a_t \sim \pi(\cdot \mid c_t; M_{k-1})$, and the rollout $\tau^{(k)}$ proceeds as described in \S\ref{sec:prelim-agent}. One or more reflector modules subsequently distill experience from the rollout and write it back into memory, $M_k = \mathrm{Reflect}(M_{k-1}, \tau^{(k)}).$ 
This forms an execute--reflect--reuse loop, enabling each task to benefit from experience accumulated from previous executions. Existing systems instantiate this loop in different ways, varying in how experience is extracted, organized, retrieved, or routed across agents \citep{legomem,smith}. 
Nevertheless, they all share the same underlying pattern: experience is abstracted from past rollouts and later reused by re-injecting it either into the model's context or into the environment.
Where these systems diverge is the form of that artifact: natural-language text or executable code. Most systems represent experience primarily as natural-language text, sometimes paired with code snippets, and differ mainly in the type of textual experience they extract, ranging from cached execution plans ~\citep{apc,memento}, evolving cross-task playbooks ~\citep{dc,ace}, and distilled reasoning strategies ~\citep{reasoningbank}, to skill documents that combine textual guidance with code examples ~\citep{skillx,xskill,procmem}.
A second family represents experience as executable code: Yunjue \citep{yunjue}incrementally grows a tool library, while Alita~\citep{alita} synthesizes reusable tools and exposes them as MCP services . Yet in all cases, the representation form is fixed \emph{a priori} at the system level rather than selected according to the characteristics of individual experiences. This representation-centric limitation is the focus of our work.

\section{The Metis System}
\label{sec:methodology}
\subsection{System Overview}
\label{sec:overview}

The overall architecture of Metis is illustrated in Figure~\ref{fig:arch}. Metis maintains a hierarchical experience memory $M=(M_{\text{text}}, M_{\text{code}})$, consisting of a structured text memory store and a code memory library. Built upon this memory hierarchy, Metis operates through a self-evolving loop comprising four key components:

\begin{figure*}[t]
  \centering
  \includegraphics[width=\textwidth]{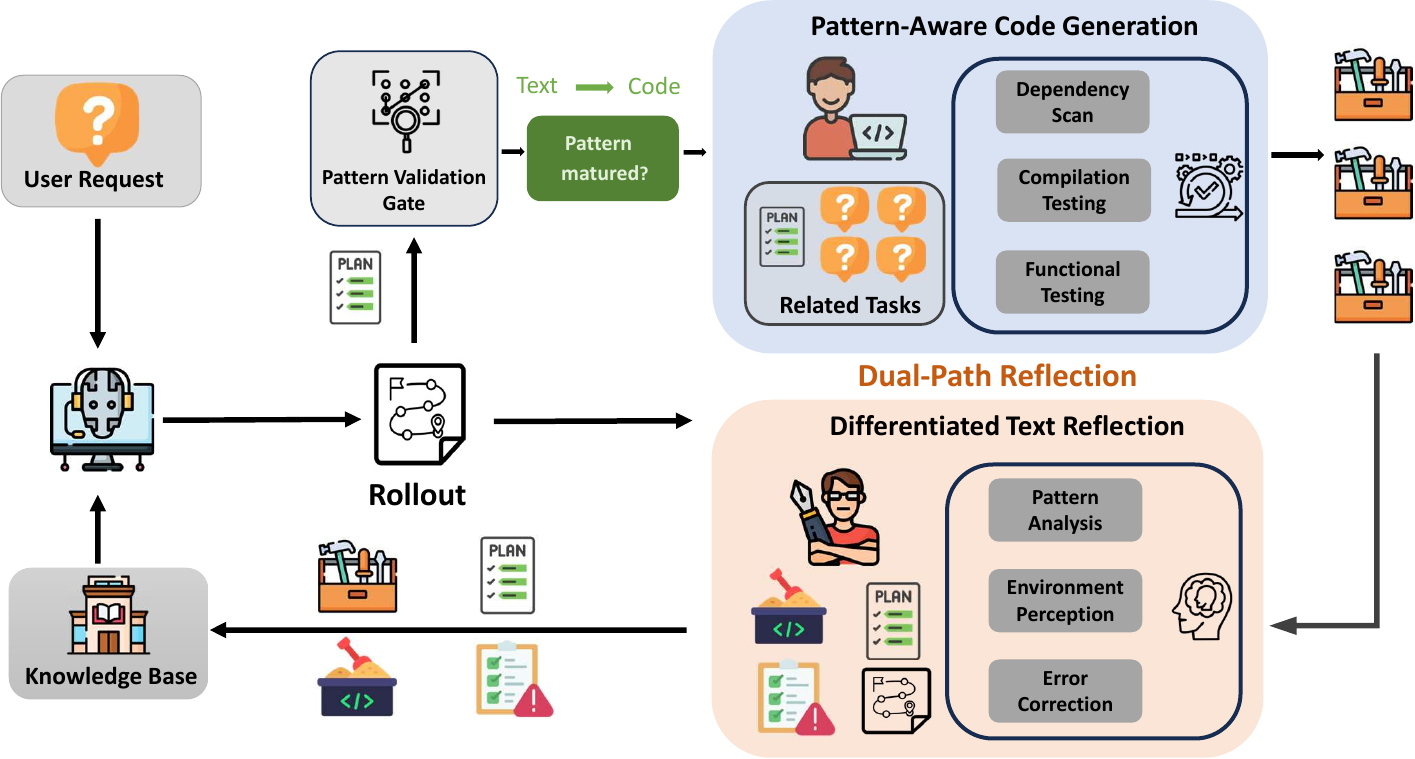}
  \caption{Architecture of Metis}
  \label{fig:arch}
\end{figure*}

\begin{enumerate}[
  label={\ding{\numexpr181+\value{enumi}\relax}},
  leftmargin=1.5em,
  labelsep=0em,
  align=left,
  itemsep=0.1\baselineskip,
  topsep=0.3\baselineskip
]

  \item \textit{Differentiated Text Reflection} distills completed trajectories into reusable textual experience. The reflector extracts three complementary forms of experience: execution plans, environment facts, and common pitfalls, corresponding to different aspects of task execution. It also maintains the text memory by merging redundant entries and replacing outdated entries with more general or more accurate versions.

  \item \textit{Pattern-Aware Code Generation} selectively promotes recurring execution plans into callable tools rather than codifying experiences indiscriminately. When a plan has been reused sufficiently often, Metis treats it as evidence of a stable behavioral pattern and invokes a codifier using the plan and related task queries, but not the raw trajectories. The generated tools are validated in the sandbox and admitted into the library only after passing dependency and compilation checks.

  \item \textit{Memory Manager} determines which memories should be reused for a new task. It first performs embedding-based top-$k$ retrieval over both text memories and tool descriptions, and then employs an LLM-based manager to jointly select the most relevant entries. Selected text memories are injected as natural-language guidance, while selected tools are compiled in the environment and exposed through their docstrings.

  \item \textit{Reflection Harness} provides controlled access to execution evidence for both text reflection and code generation. Since raw trajectories can be long and noisy, Metis exposes them through progressive disclosure. The reflector initially receives a compact trajectory view that preserves the beginning and end of the rollout while truncating the middle, and may request expansion of specific steps when finer-grained evidence is required. The harness also allows reflection agents to inspect existing tool implementations when necessary. For generated code artifacts, it further enforces dependency closure and compilation checks before admitting them into the tool library.
  
\end{enumerate}

\stitle{Workflow}
Given a user request, the \textit{Memory Manager} first selects task-relevant memories from $M$, including both textual entries and callable tools. Conditioned on the selected memories, the executor solves the task and produces an execution trajectory. After execution, Metis updates its memory through a dual-path reflection process. If the selected execution plans have accumulated enough reuse evidence, \textit{Pattern-Aware Code Generation} is triggered first to promote them into callable tools.
The \textit{Differentiated Text Reflection} then observes the executed trajectory together with the updated tool library, revising existing memories into more general and effective forms or creating new entries when necessary. 
Throughout this process, the \textit{Reflection Harness} provides controlled access to execution evidence, supports tool inspection, and enforces dependency and compilation checks. The resulting memories and tools are finally written back into $M$ for future tasks.

The remainder of this section is organized as follows. Section~\ref{sec:text_reflection} introduces differentiated text reflection and the three types of textual experience maintained by Metis. Section~\ref{sec:codify-pipeline} presents the code-generation pipeline and the recurrence-based promotion mechanism. Section~\ref{sec:manager} describes memory retrieval and selection for task execution. Finally, Section~\ref{sec:harness} introduces the reflection harness that supports both text reflection and code generation.

\subsection{Differentiated Text Reflection}
\label{sec:text_reflection}
\stitle{Memory Schema}
Metis organizes text memory into three categories,
$
M_{\text{text}} = M_{\text{env}} \cup M_{\text{pit}} \cup M_{\text{plan}}.
$
\emph{Environment facts} ($M_{\text{env}}$) capture task-independent properties of the execution environment and its constraints. 
\emph{Common pitfalls} ($M_{\text{pit}}$) describe structured failure patterns distilled from failed or inefficient executions, each represented as a \emph{(trigger, mistake, consequence)} tuple.
\emph{Execution plans} ($M_{\text{plan}}$) describe reusable templates for solving recurring sub-routines. Unlike the other two categories, execution plans capture procedural patterns and are therefore the only type of textual experience eligible for promotion into code.

Formally, text memory consists of a set of typed entries. Each entry
$
m=(\mathrm{body},\kappa,v)\in M_{\text{text}}
$
contains a natural-language body $\mathrm{body}$, a category label $\kappa\in\{\text{env},\text{pit},\text{plan}\}$, and an invalidation bit $v\in\{0,1\}$. 
An execution plan additionally carries bookkeeping fields that tie it to the code path, which we introduce together with the Codify pipeline in \S\ref{sec:codify-pipeline}. 
Text memory participates in the self-evolving loop through retrieval and reflection. For the $i$-th task, Metis first selects a task-relevant subset
$
S_i=\textsc{Select}(q_i,M_{i-1}),
$
which is injected into the executor. The executor then produces a rollout
$
\tau_i \sim \pi(\cdot \mid q_i,S_i),
$
following the interaction process defined in Section~\ref{sec:prelim-agent}.
Finally, the reflector distills experience from the rollout and updates the memory,
$
M_i=\textsc{Reflect}(M_{i-1},\tau_i).
$
The shared \textsc{Select} operator is detailed in Section~\ref{sec:manager}.

\stitle{Agentic Reflection}
After receiving the trajectory produced by the executor, the Text Memory Reflector distills it into reusable textual experience. Rather than acting as a one-shot summarizer, the reflector is itself implemented as an agent: a bounded reasoning loop with access to the same environment as the executor. This allows it to invoke the same tools and consult the same documentation, rather than relying solely on the trajectory text.

For failed tasks, the reflector probes the environment to identify the underlying cause and proposes candidate entries appropriate to the failure type, typically a corrective pitfall ($M_{\text{pit}}$) or a warning environment fact ($M_{\text{env}}$). For successful tasks, it continues to inspect the trajectory for inefficiencies, such as erroneous retries or unnecessary environment exploration, and proposes efficiency-oriented entries that would eliminate them.

To prevent execution plans from overfitting to a single task, the reflector is additionally conditioned on the queries of the most recent $N$ tasks. By observing this recent query distribution, it formulates execution plans that capture recurring patterns across tasks rather than task-specific solutions, thereby improving their generality and reusability. A candidate entry is finalized only if the reflector judges it to be sufficiently general and it passes the remaining validation gates for non-redundancy, mechanistic grounding, and correct scope.

\stitle{Memory Maintenance}
The reflector performs not only memory creation but also memory maintenance through \emph{merge} and \emph{update} operations. When multiple narrow entries are found to capture the same underlying pattern, they are merged into a single, more general entry. When a previously distilled conclusion is found to be inaccurate, biased, or superseded by newer evidence, the corresponding entry is revised accordingly.
Both operations follow an invalidate-and-replace policy rather than in-place editing. Specifically, each superseded entry has its invalidation bit flipped ($v:0\!\rightarrow\!1$), and the merged or revised entry is appended as a new record. This design preserves memory consistency while ensuring that obsolete entries are excluded from future retrieval.

\stitle{Memory Reuse}
At reuse time, the entries selected for a task are rendered as natural-language guidance and prepended to the executor's context. Because textual memories act as advice rather than executable control flow, the executor remains free to weigh, partially apply, or ignore any entry. 

This flexibility gives rise to graceful degradation: an imperfect text entry incurs at most the cost of a disregarded suggestion, whereas an executable artifact would propagate its error through every invocation. For this reason, text serves as the low-commitment default representation, while code is admitted only under the stricter conditions described in Section~\ref{sec:codify-pipeline}.

\subsection{Pattern-Aware Code Generation}
\label{sec:codify-pipeline}
\stitle{Codification Unit}
The code path turns recurring execution plans into callable tools. This promotion is selective by design: code is costly to generate and verify, and an admitted tool is executed as control flow rather than treated as optional advice. Metis therefore does not codify individual tasks or raw trajectories. Instead, it codifies only execution plans whose repeated selection indicates that the same procedural pattern recurs often enough to justify mechanization.

Execution plans are the natural unit of codification because they are the only text-memory type that specifies a mechanically executable sub-routine. Each plan $p \in M_{\mathrm{plan}}$ augments the generic text entry with three code-path fields:
\[
p = \bigl(\mathrm{body}(p), v(p), B(p), R(p), D(p)\bigr).
\]
Here $B(p)$ is the candidate buffer of recent tasks that selected $p$, $R(p)$ is a bounded related-task history retained under budget $\beta$, and $D(p)$ stores the admitted tools implementing the plan. Together, these fields make the plan the bridge between text and code: repeated selection of $p$ triggers codification, while $D(p)$ later enables the selected plan to retrieve its associated tools.

\stitle{Recurrence-Based Promotion}
Metis treats plan reuse as recurrence evidence, not as proof that the plan caused task success. Let $S_i^{\mathrm{plan}}$ denote the plans selected for task $i$. Whenever $p$ is selected, Metis updates
\[
B(p) \leftarrow B(p) \cup \{i\}, \quad \text{if } p \in S_i^{\mathrm{plan}}
\]
The task need not succeed. Even an unsuccessful rollout can indicate that the query belongs to the same procedural family. Since Codify uses the query but deliberately excludes the trajectory, such tasks contribute coverage without propagating erroneous execution paths into the generated tool. 

Codification is triggered once $|B(p)| \geq \theta$, where $\theta$ is a configurable reuse threshold. This threshold reflects the cost asymmetry between text and code: a tool is generated only when the observed recurrence suggests sufficient future reuse to amortize its construction and verification cost.
When the promotion gate fires, a codifier converts the matured plan into one or more tools. The codifier receives the plan body, the candidate queries $Q_B(p)=\{q_i \mid i\in B(p)\}$, the related queries $Q_R(p)=\{q_i \mid i\in R(p)\}$, the current tool library rendered as signatures and docstrings, and access to the sandbox environment for inspection and execution testing.

Candidate trajectories are deliberately excluded. A trajectory contains incidental variables, exploratory calls, failed attempts, and task-specific constants. While such information is useful for diagnosing a past rollout, it is unreliable evidence for constructing a reusable black-box tool. The execution plan provides the abstract procedure, the query sets define the task variations that the tool should cover, and the sandbox grounds the implementation in the actual execution environment.

\stitle{Tool Admission}
The codifier is instructed to synthesize small composable helpers rather than end-to-end solvers tailored to individual queries. Each helper must be supported by at least two structurally distinct queries from $Q_B(p)\cup Q_R(p)$, and task-varying elements such as entities, time ranges, filters, and API options must be exposed as parameters rather than hard-coded. Algorithm~\ref{alg:codify} summarizes the codification gate and state update.

\begin{algorithm}[t]
\small
\caption{Codify Pipeline}
\label{alg:codify}
\begin{algorithmic}[1]
\Require Completed task $i$; selected plans $S_i^{\mathrm{plan}}$; threshold $\theta$; history budget $\beta$; memory $M=(M_{\mathrm{text}},M_{\mathrm{code}})$
\ForAll{$p \in S_i^{\mathrm{plan}}$ with $v(p)=0$}
    \State $B(p) \leftarrow B(p) \cup \{i\}$ \Comment{accrue recurrence evidence; success not required}
    \If{$|B(p)| < \theta$} \Comment{recurrence gate: promote only matured plans}
        \State \textbf{continue}
    \EndIf
    \State $Q_B \leftarrow \{q_j \mid j \in B(p)\}$
    \State $Q_R \leftarrow \{q_j \mid j \in R(p)\}$ \Comment{queries only; trajectories deliberately excluded}
    \State $(\mathrm{ok}, T) \leftarrow \textsc{Codify}(p,Q_B,Q_R,M_{\mathrm{code}})$
    \If{\textbf{not} $\mathrm{ok}$} \Comment{failed dependency or compilation check (\S\ref{sec:harness})}
        \State \textbf{continue}
    \EndIf
    \State $M_{\mathrm{code}} \leftarrow M_{\mathrm{code}} \cup T$
    \State $D(p) \leftarrow D(p) \cup \{\mathrm{id}(t) \mid t \in T\}$ \Comment{register implementations for plan-tool linkage}
    \State $R(p) \leftarrow \mathrm{Recent}_{\beta}(R(p) \cup B(p))$ \Comment{fold consumed buffer into history}
    \State $B(p) \leftarrow \emptyset$
\EndFor
\end{algorithmic}
\end{algorithm}

A generated tool is admitted only after passing the dependency-closure and compilation checks described in Section~\ref{sec:harness}. Upon admission, the verified tools are added to $M_{\mathrm{code}}$, registered in $D(p)$, and the consumed candidate buffer is folded into the related-task history $R(p)$. Once tools become available, subsequent text reflection may revise the corresponding execution plan to reference them. In this way, recurring plans give rise to tools, while admitted tools subsequently influence future plans, closing the loop between text memory and code memory.

\subsection{Memory Manager}
\label{sec:manager}
\stitle{Joint Memory Selection}
Metis couples text and code at selection time rather than treating them as two independent memories. For task $i$, Metis first performs embedding-based top-$k$ retrieval over both text memory entries and code tools, using the query to construct a bounded candidate pool from each store. A lightweight LLM-based Memory Manager then jointly selects useful text memories and tools for the task, producing $S_i^{\mathrm{text}}$ and $S_i^{\mathrm{tool}}$. The Manager is given the task query, the retrieved text entries, and the retrieved tool signatures and docstrings, but not the full tool implementations.

Joint selection allows the Manager to reason over the interaction between textual guidance and executable artifacts. A selected plan can bring in the tools that implement it, while a selected tool can be paired with the text memory that describes when and why it should be used. The resulting selections are then passed to the dependency-checking pipeline described below. For plans, $\mathrm{Ref}(p)$ includes the registered implementation set $D(p)$, so selecting a codified plan automatically makes its tools eligible for execution.

\subsection{Reflection Harness}
\label{sec:harness}
\stitle{Dependency Closure}
After the Memory Manager selects text memories and tools, the Metis harness constructs the executable tool context through static dependency closure. We write $\mathrm{Ref}(m) \subseteq M_{\mathrm{code}}$ for the set of library tools referenced by a selected text memory $m$. For an execution plan $p$, this includes its registered implementation set $D(p)$. We further write $\mathrm{Dep}(t) \subseteq M_{\mathrm{code}}$ for the set of custom library functions called by the implementation of tool $t$. The initial executable set is
\[
U_0=S_i^{\mathrm{tool}}
\cup
\bigcup_{m \in S_i^{\mathrm{text}}}\mathrm{Ref}(m).
\]
The harness then closes this set under implementation-level dependencies:
\[
U_{r+1}=U_r
\cup
\bigcup_{t \in U_r} \mathrm{Dep}(t),
\]
and stops at the first fixed point $U_{R+1}=U_R$. The final executable tool context is
\[
U_i^{\star}=\mathrm{Close}\left(S_i^{\mathrm{text}}, S_i^{\mathrm{tool}}\right)=U_R .
\]
The resulting context contains tools directly selected for the task, tools referenced by selected text memories, and all custom helper functions required by those tools. This makes tool injection dependency-complete without requiring the Memory Manager itself to reason over implementation details.

\stitle{Tool Admission Guard}
The same dependency-closure mechanism is also applied at tool admission time. When the codifier proposes a new candidate tool $t$, the harness statically scans its implementation to identify calls to existing library tools and helper functions. The candidate is then validated in the same sandbox environment used by the executor, with all required dependencies injected through the closure procedure above. Formally, the candidate must satisfy
\[
\mathrm{Compile}\left(\mathrm{impl}(t), \mathrm{Close}(\emptyset,\{t\})\right)=1 .
\]
If compilation fails, the codifier receives the error message and is allowed one corrective round. A second failure rejects the candidate, and the tool is not added to the library. This admission guard ensures that newly codified tools are executable, dependency-complete, and compatible with the runtime environment before they can be reused.

\stitle{Progressive Disclosure}
To improve reflection efficiency, the harness exposes execution trajectories through progressive disclosure rather than placing the full trajectory directly into the reflector context. Given a trajectory
$\tau_i=\bigl(q_i,(a_1,o_1),\ldots,(a_T,o_T)\bigr)$ as defined in \S\ref{sec:prelim-agent},
the harness compresses each step independently, constructing a compact view
$\tilde{\tau}_i=(\tilde{s}_1,\ldots,\tilde{s}_T)$ with
\[
\tilde{s}_t=\bigl(\mathrm{type}(a_t),\,\mathrm{name}(a_t),\,
\mathrm{Trunc}(a_t;h),\,\mathrm{Trunc}(o_t;h)\bigr),
\]
where $\mathrm{type}(a_t)$ is the action type, $\mathrm{name}(a_t)$ the tool name when applicable, and long fields are truncated by retaining only their head and tail spans:
\[
\mathrm{Trunc}(x;h)=
\mathrm{head}_h(x) \,\Vert\, \texttt{...} \,\Vert\, \mathrm{tail}_h(x).
\]
This compact representation provides the reflector with a global view of the execution while avoiding the cost of repeatedly injecting long observations, intermediate states, and tool outputs.
The reflector is further equipped with controlled inspection tools for expanding details on demand. It may request the full content $(a_t,o_t)$ of a specific step, inspect a bounded range of steps, or view the implementation of a selected tool when code-level reasoning is required. Reflection therefore proceeds from a lightweight global sketch to targeted local inspection, rather than from a fully expanded execution trace.

\section{Experiments}
\label{sec:experiments}
\subsection{Experimental Setup}
\label{sec:experiments:setup}

\stitle{Benchmark}
We run experiments on AppWorld~\citep{appworld}, an interactive agent benchmark containing simulated environment of 9 applications exposing 457 APIs. Each task is a natural-language instruction that requires the agent to inspect the environment, compose multiple actions, and modify the underlying sandbox state to satisfy programmatic evaluation tests, mirroring deployment scenarios where an agent must perform complex operations inside a stateful environment to fulfill a user request.

\stitle{Baselines}
We compare Metis with the following baselines:
\squishlist
    \item \textsc{No Memory} runs the executor without any evolving memory and measures the performance of the base ReAct agent scaffold. We use the AppWorld's official agent implementation.

    \item \textsc{ACE}~\citep{ace} represents text-based context evolution, where past experience is maintained as natural-language guidance and retrieved into the executor context.

    \item \textsc{SkillX}~\citep{skillx} represents skill-based experience reuse, where reusable task-solving knowledge is stored as structured skill artifacts. 
\squishend
For fair comparison, we adapt all baselines to the same executor, task order, and sandbox interface, while preserving their original memory form and update logic.

\stitle{Models}
We employ Claude Sonnet 4.6 as the reflection model across all memory-based methods, as reflection quality directly bounds memory quality; since the same reflector is shared by all methods and operates only during the offline training phase, it confounds neither the comparison of memory mechanisms nor test-time cost.
For the base agent, we adopt GPT-4o with ReAct, the best-performing configuration reported in the original AppWorld benchmark~\citep{appworld}.
For methods where embedding is used, we use text-embedding-3-large as the embedding model.

\stitle{Protocol}
All methods are evaluated under a train-then-test memory-evolution protocol.
Following prior work~\citep{skillx,ace}, each method first consumes the training tasks in a single fixed-order pass to construct its experience memory, using its own construction procedure: per-rollout reflection and codification for Metis, delta-based context updates for ACE, and offline batch skill construction for SkillX.
The resulting memory is then frozen and evaluated on the held-out test tasks.
We exclude test-challenge, which is designed to assess generalization to applications absent from the training split. Since no experience-based method can accumulate relevant memory for such applications by construction, this setting evaluates out-of-domain generalization rather than experience reuse, which is not the focus of this work.

We report results under two complementary evaluation settings.
\textbf{\emph{Official}} uses the original AppWorld train and test-normal splits. These splits are disjoint at the scenario level and exhibit a noticeable distribution shift: multi-app interaction patterns are substantially more frequent in test-normal than in training. This setting therefore measures the ability of memory to generalize across unseen task compositions.
\textbf{\emph{Resampled}} pools the two splits and redraws train and test sets of the original sizes at the task level using a fixed random seed. Task-level resampling allows variants of the same scenario to appear on both sides of the split, emulating a deployment-style recurring workload where agents repeatedly encounter tasks similar to those solved previously. This is precisely the regime that experience memory is intended to support.
Together, the two settings measure complementary capabilities: generalization under distribution shift and exploitation of recurring task patterns.
Unless otherwise noted, all extra analyses are conducted under the Resampled setting: memory is exercised far more frequently in the recurring regime, so component-level differences manifest with higher resolution, whereas under distribution shift they are confounded with retrieval misses. Main results are reported under both settings.

\stitle{Metrics}
Our accuracy metric is \emph{Task Goal Completion} (TGC), defined as the fraction of tasks that pass all programmatic state-based evaluation tests.
For execution efficiency, we report executor tokens per task, measured as the total input and output tokens consumed by the base ReAct agent averaged over test tasks, together with the average number of ReAct turns per task.
For memory-construction cost, we report the total reflection tokens consumed during training. For Metis, this includes both text-reflection and codification calls, thereby accounting for all memory-construction overhead. For ACE and SkillX, we count the token cost of their reflection modules. Since different methods adopt different trajectory usage patterns, trajectory-generation cost is excluded for all methods.
We report execution cost and memory-construction cost separately because they arise at different stages. Execution cost is incurred repeatedly during inference on every task, whereas reflection cost is paid once during memory construction and amortized over future reuse.

\subsection{Main Results}
\label{sec:experiments:main}
\begin{table*}[t]
\caption{Main results on AppWorld. \emph{Official} uses the original train and test-normal splits (distribution-shifted); \emph{Resampled} redraws train/test of the original sizes from the pooled tasks (recurring workload). Execution Tokens and ReAct Turns are averaged per test task; Reflection Cost is the total one-time cost incurred during the training phase and is amortized offline.}
\centering
\small
\setlength{\tabcolsep}{8pt}
\begin{tabular}{lcccc}
\toprule
Method & Acc. (\%) $\uparrow$ & Execution Tokens (K) $\downarrow$ & ReAct Turns $\downarrow$ & Reflection Cost (M) \\
\midrule
\multicolumn{5}{c}{\textit{Official split}} \\
\midrule
No Memory             & 51.8          & 112.6          & 14.55          & 0    \\
ACE                   & 53.6          & 422.4          & 13.60          & 6.9 \\
SkillX                & 54.8          & 100.3          & 12.30          & 9.4  \\
\textbf{Metis (Ours)} & \textbf{60.1} & \textbf{97.4}  & \textbf{11.25} & 7.9  \\
\midrule
\multicolumn{5}{c}{\textit{Resampled split}} \\
\midrule
No Memory             & 54.8          & 101.7          & 13.92          & 0    \\
ACE                   & 57.7          & 204.8          & 11.90           & 4.3   \\
SkillX                & 65.5          & 89.5           & 11.03          & 17.8 \\
\textbf{Metis (Ours)} & \textbf{66.1} & \textbf{78.5}  & \textbf{10.32} & 11.8 \\
\bottomrule
\end{tabular}
\label{tab:main}
\end{table*}
\stitle{Overall Comparison}
The results are summarized in Table~\ref{tab:main}. 
Across both splits, Metis achieves the highest task accuracy while consuming the fewest executor tokens and ReAct turns, improving TGC over the no-memory baseline by 8.3 points on the Official split and 11.3 points on the Resampled split, where ACE and SkillX gain at most 2.9 and 10.7 points respectively.
Notably, the gains in accuracy do not come at the price of efficiency: Metis reduces executor tokens by 13.5\% and 22.8\% relative to the baseline on the two splits, whereas ACE inflates execution cost by up to 3.8x despite its smaller accuracy gain.
Metis also incurs lower reflection cost than SkillX on both splits, even though it additionally performs tool synthesis and validation.
Metis is thus the only method that improves accuracy and execution efficiency simultaneously while spending less reflection budget than the strongest baseline.
The following paragraphs attribute each improvement to its corresponding design decision.

\stitle{Robustness under Distribution Shift}
The two splits differ sharply in how far test tasks deviate from the training distribution, and the accuracy gaps widen accordingly; since ACE yields only marginal accuracy gains on both splits, we focus here on Metis and SkillX.
On the Resampled split, where test tasks recur from scenario families seen during training, SkillX performs comparably to Metis (65.5 vs. 66.1): when relevant experience is directly available, most reasonable memory designs can exploit it.
On the Official split, all methods decline, but the no-memory baseline isolates how much of the decline reflects intrinsic task difficulty: the baseline drops 3.0 points, while the gain over it shrinks from +10.7 to +3.0 for SkillX yet only from +11.3 to +8.3 for Metis.
That is, once task difficulty is controlled, SkillX loses the majority of its memory utility under distribution shift, whereas Metis retains most of it.

The divergence stems from what each system chooses to crystallize. SkillX constructs skills through offline synthesis without recurrence evidence, causing the resulting artifacts to retain assumptions specific to the tasks from which they were distilled. In contrast, Metis promotes an execution plan only after it has been repeatedly selected across distinct queries and requires task-varying elements to be exposed as parameters. As a result, the generated tools are distilled from a broader set of observations and are explicitly designed to accommodate task variation, yielding better transferability.

Experiences that do not satisfy the promotion criteria remain in textual form, which transfers more reliably across tasks, as discussed in Section~\ref{sec:introduction}. Under distribution shift, this selectivity separates experience that captures reusable procedural structure from experience that offers little opportunity for future reuse.

\begin{table}[t]
\begin{minipage}[t]{0.46\textwidth}
\centering
\small
\caption{Execution cost on the commonly solved tasks.}
\label{tab:common-cost}
\begin{tabular}{lcc}
\toprule
Method & Tokens (K) $\downarrow$ & Turns $\downarrow$ \\
\midrule
No Memory & 85.2 & 12.2 \\
SkillX & 71.4 & 9.7 \\
\textbf{Metis (Ours)} & \textbf{59.6} & \textbf{8.6} \\
\bottomrule
\end{tabular}
\end{minipage}
\hfill
\begin{minipage}[t]{0.50\textwidth}
\centering
\small
\caption{Knowledge reuse on the test split. $^*$Injected at least once. $^+$Invoked at least once.}
\label{tab:reuse}
\begin{tabular}{lcc}
\toprule
Metrics & Metis & SkillX \\
\midrule
Tasks w/o memory injection $\downarrow$ & 0\% & 20.8\% \\
Plan-level entries reused$^*$ $\uparrow$ & 83\% & 86\% \\
Code-bearing entries reused$^+$ $\uparrow$ & 94\% & 84.1\% \\
\bottomrule
\end{tabular}
\end{minipage}
\end{table}
\stitle{Execution Efficiency}
Aggregate per-task cost in Table~\ref{tab:main} reflects both execution efficiency and task coverage: methods that solve different subsets of tasks may incur different costs simply because the solved tasks differ in difficulty. To control for this effect, we additionally compare execution cost on the \emph{commonly solved} subset, consisting of the 62 tasks successfully completed by No Memory, SkillX, and Metis alike. Results are shown in Table~\ref{tab:common-cost}.
We exclude ACE from this controlled comparison: its playbook-injection cost grows with accumulated memory regardless of the task at hand, so its per-task figures are dominated by memory size rather than solving behavior, as Table 1 already shows.
On this subset, Metis requires 59.6K tokens and 8.6 turns per task, compared with 71.4K/9.7 for SkillX and 85.2K/12.2 for No Memory. The difference stems from how experience is consumed. 
SkillX retrieves relevant skills selectively, but even its code-bearing functional skills are presented as contextual information that the agent must read and reason over at inference time. Metis, by contrast, registers recurring routines as callable tools, allowing the agent to invoke them directly rather than reconstructing the procedure through reasoning or code generation. This difference in consumption form leads to substantially higher execution efficiency.

\stitle{Construction Cost and Amortization}
Reflection cost yields different returns across systems. ACE achieves a low memory-construction cost through its delta-update strategy, but as shown above, this investment translates into little execution-time benefit. Both Metis and SkillX convert reflection effort into execution savings; however, Metis achieves larger savings at lower construction cost despite additionally synthesizing and validating tools.
This advantage stems from the recurrence-based promotion mechanism. Metis codifies only plans that have accumulated sufficient reuse evidence, whereas SkillX synthesizes skills directly from individual task experiences. As shown in our profiling study in Section~\ref{sec:introduction}, tool generation is substantially more expensive than text reflection and is worthwhile only when the resulting artifact is reused repeatedly.

\begin{table*}[t]
\caption{Ablations on the AppWorld dev split. Construction Cost: total memory-construction tokens; Execution Tokens and ReAct Turns are averaged per task. Text-only and Code-only each retain a single representation form; Eager removes the recurrence gate and codifies after every task.}
\centering
\small
\begin{tabular*}{\textwidth}{@{\extracolsep{\fill}}lcccc@{}}
\toprule
Variant & Acc. (\%) $\uparrow$ & Construction Cost (M) $\downarrow$ & Execution Tokens (K) $\downarrow$ & ReAct Turns $\downarrow$ \\
\midrule
No Memory          & 56.1 & 0    & 74.4 & 11.46 \\
Metis (Text-only)  & 62.3   & 4.96   & 55.6   & 9.25    \\
Metis (Code-only)  & 60.2   & 4.89   & 58.7   & 8.89    \\
Metis (Eager)      & 63.2 & 11.6 & 49.9 & 7.68  \\
\midrule
\textbf{Metis (full)} & \textbf{66.7} & 7.9 & 54.5 & 8.46 \\
\bottomrule
\end{tabular*}
\label{tab:ablation}
\end{table*}

To examine whether the constructed knowledge is actually exercised, we compare reuse patterns in Table~\ref{tab:reuse}. At the task level, every Metis test task receives at least one relevant memory, whereas 20.8\% of SkillX's test tasks retrieve nothing, leaving the agent to solve them from scratch.
At the library level, plan-level reuse is similar between the two systems (83\% vs.\ 86\%), but the difference emerges for code-bearing artifacts. Among admitted tools, 94\% of Metis's tools are \emph{invoked} by the executor at least once, whereas only 84.1\% of SkillX's code-bearing skills are \emph{injected} into the context.
This comparison is \textbf{conservative} because invocation is a strictly stronger criterion than injection. An injected skill may be presented to the agent yet never influence its behavior, whereas an invoked tool necessarily participates in execution. The near-universal use of admitted tools suggests that the recurrence gate successfully identifies patterns with genuine future reuse, ensuring that construction cost is spent predominantly on knowledge that is later exercised.

\subsection{Ablations}
To disentangle the effects of individual design decisions, we conduct ablation studies to answer two questions:
\squishlist
    \item Does the hybrid representation outperform either representation form alone?
    \item Does recurrence-gated promotion outperform eager per-task codification?
\squishend
To reduce evaluation cost, all ablations construct memory on the official training split and evaluate on the dev split, following the benchmark's standard protocol~\citep{appworld}. Each variant rebuilds its memory from scratch under its own configuration rather than modifying or masking the memory produced by the full system. Consequently, every row in Table~\ref{tab:ablation} reflects the end-to-end behavior of the corresponding design. Reconstruction is necessary because text memories and code tools are co-evolved in Metis. Reusing memory generated by the full system would leak information across variants and would be unfair to the single-representation baselines.

\stitle{Memory Forms}
To isolate the contribution of each representation, we evaluate two single-form variants: \emph{Text-only}, which disables the code path entirely, and \emph{Code-only}, which disables text memory. Note that without execution plans, the recurrence gate has no signal to operate on, so Code-only necessarily codifies per task; this variant thus reflects the ungated codification paradigm, while the gate's independent effect is isolated in the next ablation analysis.

As Table~\ref{tab:ablation} shows, both single forms improve over No Memory (+6.2 points for Text-only, +4.1 for Code-only), yet the full system exceeds them by a further 4.4 and 6.5 points respectively. The hybrid gain is therefore not additive bookkeeping but interaction: plans supply the recurrence signal that decides what deserves codification, and admitted tools in turn sharpen subsequent plans, a loop neither form sustains alone.

Notably, Code-only remains above the no-memory baseline rather than collapsing below it as per-task code reflection did in our profiling study (\S\ref{app:experiments}). The reconciliation lies in what the two codifiers consume and what guards their output. First, the Metis codifier is query-only by design: unlike the profiling reflector, it never reads trajectories, closing the channel through which faulty execution paths were crystallized into tools. Second, the ablation variants inherit the full Metis harness, including sandbox validation, dependency closure, and compilation-gated admission, protections absent from the deliberately minimal profiling setup. The construction-cost asymmetry persists in interaction depth rather than tokens: the text reflector averages 2.84 reflection turns per task against the codifier's 6.08, while query-only prompting keeps the token totals comparable (4.96M vs. 4.89M).

\stitle{Promotion Strategies}
We further compare recurrence-gated promotion against an \emph{Eager} variant that keeps the hybrid representation but triggers codification immediately after every task, removing the recurrence gate while holding all other components fixed. Eager codification incurs 47\% higher construction cost (11.6M vs. 7.9M tokens) and degrades dev accuracy by 3.5 points (63.2 vs. 66.7). Notably, its larger tool library does yield marginally lower execution cost (49.9K vs. 54.5K tokens per task; 7.68 vs. 8.46 turns), but this saving is bought with disproportionate construction spending and an accuracy loss: tools distilled without recurrence evidence overfit to individual tasks and transfer poorly. This is an attenuated form of the per-task codification failure in our profiling study, softened here by the query-only codifier and harness safeguards. Utilization statistics corroborate the waste: only 41\% of Eager's tools are ever invoked, against 56\% for the full system. We note that the absolute rates are bounded by the small dev split, which simply offers too few tasks to exercise every admitted tool; on the larger test split, Metis's invocation rate reaches 94\% as shown in Table~\ref{tab:reuse}, confirming that its gated tools do get exercised given sufficient workload. The 15-point gap on the identical dev workload therefore isolates the quality difference: Eager's surplus tools are not awaiting their tasks but overfit to tasks already past. The gate, in other words, is not free efficiency but a quality filter: it concedes a sliver of execution saving to avoid paying construction cost for tools that do not generalize.

\section{Conclusion}
\label{sec:conclusion}
In this work, we revisit a design dimension that existing self-evolving agents fix a priori: the representation form of accumulated experience. Our controlled study shows that text and code memories are complementary rather than interchangeable, motivating Metis, which maintains differentiated text memory as a low-commitment foundation and selectively promotes recurring plans into validated callable tools. 
Experiments on AppWorld demonstrate that Metis achieves the highest task accuracy at the lowest execution cost among the compared methods, while spending less reflection budget than the closest competitor despite additionally synthesizing and validating tools; our ablations attribute these gains to the hybrid representation and the recurrence gate respectively.
 
\bibliographystyle{iclr2026_conference}
\bibliography{main}

\appendix
\section{Experiments}
\label{app:experiments}

\subsection{Profiling Experiments}
\label{app:experiments:profiling}

This appendix reports the full protocol and the per-axis analysis behind the profiling study previewed in \S\ref{sec:introduction}. The study isolates the two experience forms over an identical stream of executions, and characterizes each along the three axes of Figure~\ref{fig:trade-offs}: \emph{construction cost}, \emph{execution efficiency}, and \emph{transfer reliability}. The goal is to reveal the underlying features of each form and the trade-offs they bear to inspire how to design a more effective and efficient self-evolving agents.

\subsubsection{Experimental Setup}
\label{app:experiments:setup}

\paragraph{Benchmark and roles.}
All profiling runs are on the AppWorld~\citep{appworld} \texttt{train} split (90 tasks). We fix a strong, widely-used executor, GPT-4o, and a separate, more capable reflector, Claude-Sonnet-4.6; keeping the two roles on different models isolates the cost of \emph{constructing} knowledge (reflector) from the cost of \emph{using} it (executor), and prevents a single model's idiosyncrasies from confounding both sides at once.

\paragraph{Execution Protocol.}

Unless stated otherwise, knowledge evolves in a streaming regime that mirrors deployment. Tasks are processed in a fixed order; after each task the reflector distills knowledge from that task's trajectory and commits it to memory, and the executor may retrieve and apply that memory on every subsequent task. Each task is therefore evaluated against a memory built only from strictly earlier tasks. To analyze the transfer reliability, we further conducted an extra experiment under oracle setting. Under this setting, we (i)~first run all 90 tasks to collect their trajectories, (ii)~reflect on each trajectory independently and in batch, so no online ordering effect or cross-task contamination is in play, (iii)~assemble the full tool library, (iv)~manually remove the single tool we identified as broken (\emph{all other tools left intact, with no further edits}), and (v)~re-run all 90 \texttt{train} tasks with that library injected. Two properties make this setting an upper bound rather than a generalization measure: the library is built from the entire task set and then hand-audited, and it is re-tested in-sample on the very tasks it was distilled from. We apply the same batch-reflection protocol to Text (without manual modification) to obtain its Oracle point for comparison.

\paragraph{Conditions.}
We compare three conditions that share this protocol and differ only in the form in which experience is stored: \textbf{No Memory} (the knowledge-free baseline; the executor sees neither tips nor tools), \textbf{Text} (the reflector distills experience as natural-language text memories), and \textbf{Code} (the reflector compiles recurring procedures into callable tools). A consequence of the two forms is an injection asymmetry that the analysis turns on: a text tip is injected directly into the executor's context, so the executor reads its full content and decides how to act on it; a code tool is compiled into a callable object, so the executor sees only its docstring and signature and invokes it as a black box, inheriting whatever its body does.

\paragraph{Metrics.}
We report task accuracy (task goal completion, TGC), the execution cost on the executor side (total tokens and ReAct steps), and the construction cost on the reflector side (total reflection tokens and LLM rounds). For execution efficiency we report both All Tasks and the Commonly Solved subset---the tasks solved by all three conditions. Commonly Solved is the apples-to-apples comparison: it removes the confound that the conditions succeed on different task sets, so a token or step difference reflects how each form solves the same work rather than which work each happened to finish.

\subsubsection{Per-Axis Analysis}
\label{app:experiments:analysis}

\begin{figure}[t]
  \centering
  \begin{subfigure}{0.45\linewidth}
    \includegraphics[width=\linewidth]{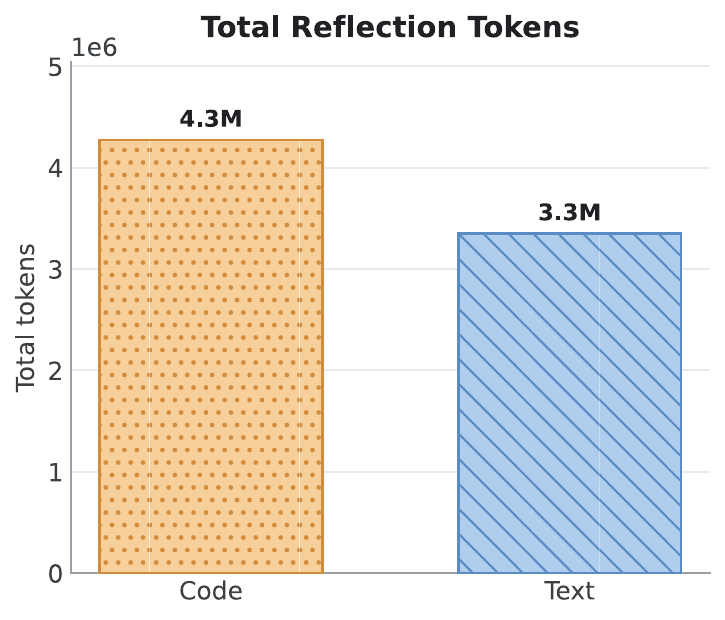}
    \caption{Reflection tokens}
  \end{subfigure}
  \hfill
  \begin{subfigure}{0.45\linewidth}
    \includegraphics[width=\linewidth]{images/fig5_reflect_steps.pdf}
    \caption{Reflection rounds}
  \end{subfigure}
  \caption{Construction cost over the streaming run. Building code costs $1.3\times$ the reflection tokens but $2.5\times$ the LLM rounds of building text,
  because code must be tested and debugged in a live sandbox before a tool is admitted.}
  \label{fig:prof-cost}
\end{figure}

\paragraph{Construction cost: text is cheaper to build.}
Over the streaming run, code reflection consumes 4.3M tokens across 562 LLM rounds, against text's 3.3M tokens across 224 rounds
(Figure~\ref{fig:prof-cost}). The round count is the sharper signal: building code costs $2.5\times$ the LLM rounds of building text, while the token gap is a
more modest $1.3\times$. The disparity follows from \emph{how} each form is produced. Text is distilled in essentially a single read of the trajectory, so each
round is one long prompt; code is produced interactively---the reflector drafts a candidate tool and must then inspect API documentation, execute the tool in a
live sandbox, and debug it before the tool is admitted, which multiplies rounds while keeping each round short. We are deliberate about not overstating the
token gap: per unit of memory, code is only moderately more expensive in raw tokens. The cost that matters is that code's heavier construction loop (sandbox
execution, verification, retries) is, as the transfer reliability axis shows, \emph{largely wasted under streaming}---the tools it pays to build do not transfer.

\begin{figure}[t]
  \centering
  \begin{subfigure}{0.49\linewidth}
    \includegraphics[width=\linewidth]{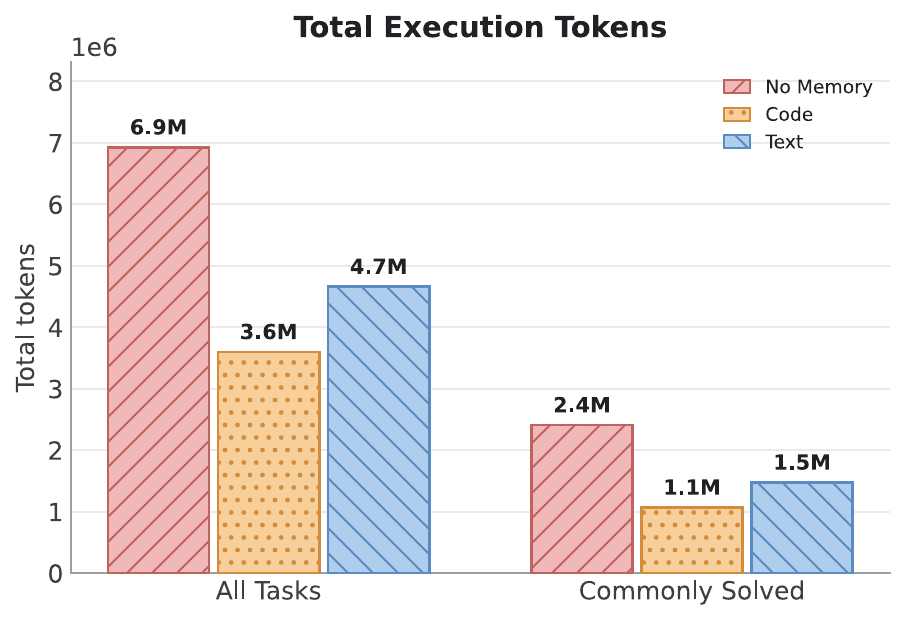}
    \caption{Execution tokens}
  \end{subfigure}
  \hfill
  \begin{subfigure}{0.49\linewidth}
    \includegraphics[width=\linewidth]{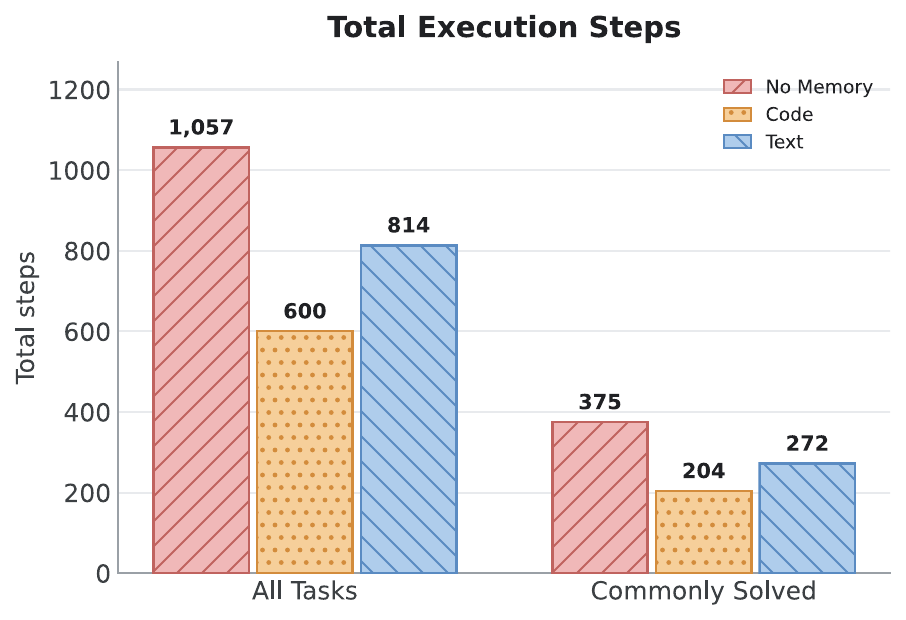}
    \caption{Execution steps}
  \end{subfigure}
  \caption{Execution efficiency. On the Commonly Solved subset (same tasks, all three conditions succeed), code cuts execution tokens and steps substantially
  more than text relative to the no-memory baseline.}
  \label{fig:prof-efficiency}
\end{figure}

\paragraph{Execution efficiency: code accelerates execution far more.}
On the Commonly Solved subset---the fair, same-tasks comparison---code drives the executor to 1.1M tokens and 204 steps, text to 1.5M tokens and 272 steps,
against the no-memory baseline's 2.4M tokens and 375 steps (Figure~\ref{fig:prof-efficiency}). Both forms accelerate execution, but code's reduction is the
larger by a wide margin: relative to baseline it cuts tokens by $54\%$ and steps by $46\%$, where text cuts them by $38\%$ and $27\%$---roughly $1.4$--$1.7\times$
the per-task saving, and head-to-head code is about a quarter leaner than text. The mechanism is the injection asymmetry. A callable tool collapses an entire
procedure---API discovery, composition, and boilerplate---into one invocation that \emph{bypasses reasoning}, whereas a text tip must be read and reasoned over
on \emph{every} use: the agent still re-derives the procedure each time, only faster and more reliably. This is precisely why text yields little efficiency gain
relative to code even though it does shorten executions versus the baseline.

\begin{figure}[t]
  \centering
  \begin{subfigure}{0.49\linewidth}
    \includegraphics[width=\linewidth]{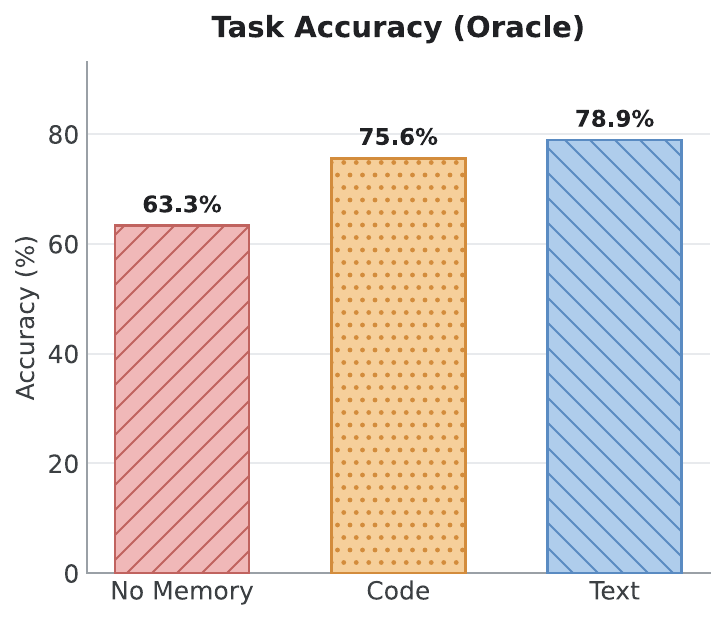}
    \caption{Oracle}
  \end{subfigure}
  \hfill
  \begin{subfigure}{0.49\linewidth}
    \includegraphics[width=\linewidth]{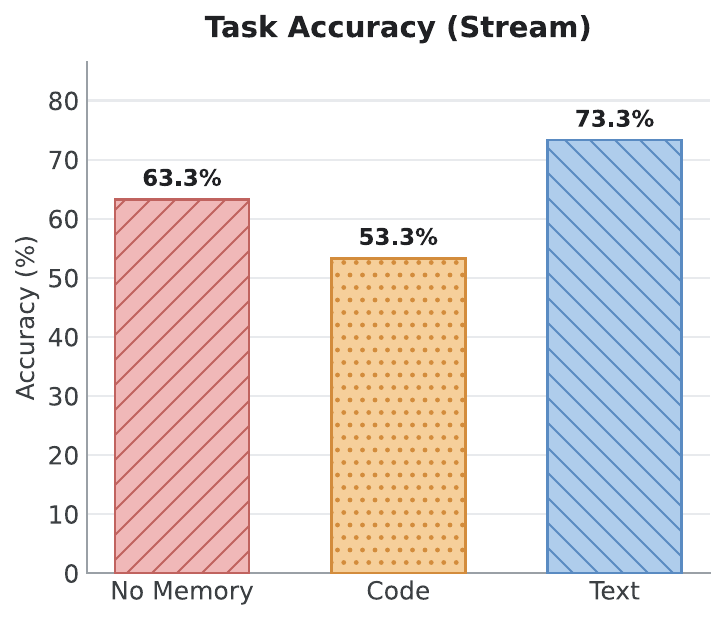}
    \caption{Streaming}
  \end{subfigure}
  \caption{Transfer Reliability. Under the in-sample Oracle, code nearly matches text; under streaming forward transfer, code collapses by 22.3 points to \emph{below}
  the no-memory baseline, while text degrades by only 5.6 points and stays well above it.}
  \label{fig:prof-reuse}
\end{figure}

\paragraph{Text memory has better transfer reliability.}
\label{app:experiments:reusability}
The third axis is the subtle one, and it has two facets. The first is \emph{generalization}, read directly off the Oracle--Streaming contrast
(Figure~\ref{fig:prof-reuse}). Under the Oracle---an in-sample upper bound---code reaches $75.6\%$, $+12.3$ points over the $63.3\%$ baseline and close to
text's $78.9\%$. Under streaming forward transfer, code \emph{collapses} to $53.3\%$, $10.0$ points \emph{below} the no-memory baseline, while text holds at
$73.3\%$, still $+10.0$ above it. Code's Oracle$\rightarrow$Streaming drop is thus $22.3$ points against text's $5.6$. The reading is unambiguous: code's strong
Oracle number is in-sample---the library is built from and tested on the same tasks and hand-cleaned---so it is an optimistic ceiling, and the gap between that
ceiling and the streaming measurement is exactly code's failure to transfer to tasks unseen at build time. Per-task code reflection \emph{overfits to the
single trajectory it was distilled from}; text transfers with only minor degradation.

The second facet, which explains the first, is \emph{safety}, and it again turns on the injection asymmetry. Reflecting over one trajectory, the reflector tends to crystallize tools that are over-specialized to that task---and sometimes generalized from an execution path that was itself \emph{wrong}. Text reflection is
prone to the same over-fitting, but the consequence differs sharply because of how each form is consumed. A text tip is in-context advice: the executor reads it and applies it with discretion, adapting it, partially applying it, or ignoring it when it does not fit, and it continues to sample the environment and self-correct. A code tool is a black box that the executor calls verbatim, so a tool distilled from a faulty trajectory propagates its defect to every caller;
we further observe that the mere presence of a confident-looking tool \emph{suppresses} the agent's own recovery behavior---it trusts the tool's (possibly empty
or wrong) return rather than re-checking the environment, a recovery loop that text-only runs routinely exercise. Text is therefore the safer carrier of freshly-distilled, not-yet-verified experience; code becomes safe only once a pattern has been verified---which is exactly the privilege the Oracle condition grants and that streaming code never earns.

\paragraph{From profiling to Metis.}
The two designs at the core of Metis are direct responses to the two failure modes this study exposes. The transfer reliability analysis shows that text is the right
default carrier for cold-start, judgment-bearing, and not-yet-verified experience---but only if the reflector knows \emph{what kind} of experience it is
distilling and phrases it for safe, adaptive reuse. This is why Metis does not treat text as a flat playbook but organizes it by experience type, a taxonomy
(environment facts, execution plans, execution pitfalls) we arrived at precisely while diagnosing which text distillations transferred and which
misled---yielding \emph{Differentiated Text Memory} (\S\ref{sec:text_reflection}). The efficiency and cost axes show that code pays off handsomely once
a pattern has genuinely recurred and been verified, yet firing code reflection per task is what produced both the streaming overfitting and the wasted
construction cost. Metis therefore gates codification on the text memory and on observed recurrence, and validates each tool before admission---\emph{Pattern-Aware
Code Generation} (\S\ref{sec:codify-pipeline})---so the high construction cost of code is paid only for patterns that have already proven worth crystallizing.

\end{document}